\documentclass[11pt,a4paper]{article}
\usepackage[hyperref]{emnlp2020}
\usepackage{times}
\usepackage{latexsym}
\usepackage{booktabs}
\usepackage[T1]{fontenc}
\usepackage{amssymb}
\usepackage{pifont}
\usepackage{url}
\usepackage{multirow}
\usepackage{graphicx}
\usepackage{bm}
\usepackage{subfigure}

\usepackage{microtype}

\aclfinalcopy 
\def\aclpaperid{252} 

\title{\textsc{Union}: An Unreferenced Metric for Evaluating Open-ended Story Generation}

\author{
 Jian Guan, Minlie Huang\Thanks{~Corresponding author}\\
Department of Computer Science and Technology, Institute for Artificial Intelligence,\\
State Key Lab of Intelligent Technology and Systems,\\
Beijing National Research Center for Information Science and Technology,\\
Tsinghua University, Beijing 100084, China\\
  \texttt{j-guan19@mails.tsinghua.edu.cn}, \texttt{aihuang@tsinghua.edu.cn} \\
}
\date{}

\begin{document}
\maketitle
\begin{abstract}
Despite the success of existing referenced metrics~(e.g., BLEU and MoverScore), they correlate poorly with human judgments for open-ended text generation including story or dialog generation because of the notorious one-to-many issue: there are many plausible outputs for the same input, which may differ substantially in literal or semantics from the limited number of given references. 
To alleviate this issue, we propose {\textsc{Union}}, a learnable \textit{\textbf{UN}referenced metr\textbf{I}c for evaluating \textbf{O}pen-e\textbf{N}ded story generation}, which measures the quality of a generated story without any reference. Built on top of BERT, \textsc{Union} is trained to distinguish human-written stories from negative samples and recover the perturbation in negative stories. We propose an approach of constructing negative samples by mimicking the errors commonly observed in existing NLG models, including repeated plots, conflicting logic, and long-range incoherence. 
Experiments on two story datasets demonstrate that \textsc{Union} is a reliable measure for evaluating the quality of generated stories, which correlates better with human judgments and is more generalizable than existing state-of-the-art metrics.
\end{abstract}

\begin{table}[!ht]
\small
    \centering
    \begin{tabular}{p{7.2cm}}
    \toprule
    {\textbf{Leading Context}}\\
    
    {Jack was at the bar. }\\
    \midrule
    {\textbf{Reference By Human}}\\
    {He noticed a phone on the floor. He was going to take it to lost and found. But it started ringing on the way. Jack answered it and returned it to the owner's friends.} \\
\midrule
{\textbf{Sample 1~(Reasonable, B=0.29, M=0.49, U=1.00)}}\\
 {On the way out he noticed a phone on the floor. He asked around if anybody owned it. Eventually he gave it to the bartender. They put it into their lost and found box.} \\\\

{\textbf{Sample 2~(Reasonable, 
B=0.14, M=0.27, U=1.00)}}\\  
{He had a drinking problem. He kept having more beers. After a while he passed out. When he waked up, he was surprised to find that he lost over a hundred dollars.}\\\\

{\textbf{Sample 3~(Unreasonable, 
B=0.20, M=0.35, U=0.00)}}\\ 
He was going to get drunk and get drunk. The bartender told him it was already time to leave. Jack started drinking. Jack wound up returning but cops came on the way home. \\
		\bottomrule
    \end{tabular}
    \caption{Generated story samples given the same leading context from ROCStories~\cite{mostafazadeh2016corpus}.  \textbf{B} stands for BLEU~\cite{papineni2002bleu},  \textbf{M} for  MoverScore~\cite{zhao2019moverscore}, and \textbf{U} for \textsc{Union}. A story can be reasonable even if it is dissimilar to the reference with a low BLEU score (B=0.14 in Sample 2), or unreasonable even if it has a large MoverScore (M=0.35 in Sample 3). In contrast,  \textsc{Union} is more reliable for evaluating story generation.
    }
    \label{tab:example}
\end{table}

\section{Introduction}
Significant advances have been witnessed with neural encoder-decoder paradigm~\cite{sutskever2014sequence}, transformer-based architecture~\cite{vaswani2017attention} and large-scale pretraining models~\cite{devlin2018bert,radford2019language} in a wide array of natural language generation~(NLG) tasks including machine translation~\cite{bahdanau2015neural}, 
story generation~\cite{fan2018hierarchical,guan2020knowledge}, and many more. 
However, the research is increasingly hindered by the lack of effective evaluation metrics, particularly for open-ended text generation tasks such as  
story generation. 

Since human evaluation is time-consuming, expensive, and difficult to reproduce, the community commonly uses automatic metrics for evaluation. Previous studies in conditional language generation tasks~(e.g., machine translation) have developed several successful referenced metrics, which roughly quantify the lexical overlap~(e.g., BLEU~\cite{papineni2002bleu}) or semantic entailment (e.g., MoverScore~\cite{zhao2019moverscore})  between a generated sample and the reference. However, such referenced metrics correlate poorly with human judgments when evaluating open-ended text generation~\cite{liu2016not} due to the one-to-many nature~\cite{zhao2017learning}, as illustrated in  Table~\ref{tab:example}. 
Specifically, 
a generated sample can be reasonable if it is coherent to the given input, and self-consistent within its own context 
but not necessarily being similar to the reference in literal or semantics, as shown in Sample 2 and 3. 

To address the one-to-many issue,  unreferenced metrics are proposed to measure the quality of a generated sample without any reference. \citeauthor{kannan2017adversarial}~(\citeyear{kannan2017adversarial}) presented a learnable, unreferenced metric which measures the text quality by learning to distinguish human-written texts from generated samples. However, the discriminator-based metric can easily lead to over-fitting to specific data~\cite{garbacea2019judge} or model bias since the quality of generated texts varies substantially across different NLG models.
As a matter of fact, the \textit{generalization or robustness} issue is critical for any learnable metrics.

Therefore, 
we propose \textsc{Union}, a learnable \textit{\textbf{UN}referenced metr\textbf{I}c for evaluating \textbf{O}pen-e\textbf{N}ded story generation}.  
\textsc{Union} learns 
to distinguish human-written stories from negative samples auto-constructed by generating perturbations of human-written stories. 
It is trained without dependence on specific NLG models or any human annotation,
making it more generalizable to distribution drift \cite{sellam2020bleurt} than the discriminator-based metric and those metrics which learn from human preference~(e.g., Adem~\cite{lowe2017towards}). To capture commonly observed issues in generated stories, such as repeated plots, conflicting logic, and inter-sentence incoherence, we adopt four negative sampling techniques to construct negative samples, including repetition, substitution, reordering, and negation alteration. 
In addition, we design an auxiliary reconstruction objective for \textsc{Union}, which recovers the perturbation from a negative sample.  This objective is shown to further improve the performance of \textsc{Union}.

Our contributions are summarized as follows:
\textbf{I.} We propose a learnable unreferenced metric \textsc{Union} for evaluating open-ended story generation to alleviate the one-to-many issue of referenced metrics. \textsc{Union} does not depend on any output of NLG models or human annotation. 
\\
\textbf{II.} Extensive experiments\footnote{All the codes and data are available at \url{https://github.com/thu-coai/UNION}.} show that \textsc{Union} correlates better with human judgments than state-of-the-art metrics, and is more generalizable to data drift (samples from different datasets) and quality drift (samples with different quality levels).




\section{Related Work}
Automatic evaluation is crucial for language generation tasks. We roughly divide existing metrics into referenced, unreferenced, and hybrid metrics, according to whether they rely on human-written references when calculating the metric score. 

\noindent\textbf{Referenced metrics} usually measure how similar a generated text is to the reference text. Therefore, they are developed mainly for conditional language generation tasks such as machine translation and text summarization, where plausible outputs are largely limited within the semantics of input. Commonly used referenced metrics include word-overlap based~(e.g., BLEU~\cite{papineni2002bleu}, ROUGE~\cite{lin-2004-rouge}) and embedding based metrics~(e.g., BertScore~\cite{zhang2019bertscore}, MoverScore~\cite{zhao2019moverscore}). 
However, 
referenced metrics are reported to correlate poorly with human judgments in open-ended generation tasks including open-domain dialog generation~\cite{liu2016not} and story generation, where the input contains only limited information for generation, and there are many plausible outputs for the same input, which can vary substantially in literal or semantics.


\noindent\textbf{Unreferenced metrics} measure the quality of a sample without any reference. The most classic unreferenced metric is \textit{perplexity}
, which measures how likely a sample is generated by a given language model trained on human-written texts. However, recent work has shown that natural language is rarely the most probable text~\cite{holtzman2019curious}, and perplexity is inadequate to measure quality~\cite{hashimoto2019unifying}. Therefore, perplexity may not indicate the actual text quality well. 
Discriminator-based metric~\cite{kannan2017adversarial} measures how easily a discriminator distinguishes the generated samples from human-written texts. However, training such a discriminator can be easily over-fitted to a specific dataset, thereby leading to poor generalization and low correlation with human judgments~\cite{garbacea2019judge}. In addition to the above point-wise metrics which score an individual sample, \citeauthor{semeniuta2018accurate}~(\citeyear{semeniuta2018accurate}) proposed the Fr\'echet InferSent Distance (FID) to evaluate the model-level quality and diversity of generated samples, 
by computing the Fr\'echet distance between the Gaussian distribution fitted to human text embeddings and that to generated sample embeddings. However, in real data, the distribution of embeddings may be far from Gaussian. 
Recently, \citeauthor{zhou2020learning}~(\citeyear{zhou2020learning}) proposed to 
evaluate sample-level quality by comparing 
a pair of samples, and further adopted a skill rating system to evaluate model-level quality based on the sample-level pair-wise comparison. However, it is unlikely to evaluate a single sample without access to its references.

\noindent\textbf{Hybrid metrics} combine referenced and unreferenced metrics. 
For open-domain dialog system evaluation, \citeauthor{lowe2017towards}~(\citeyear{lowe2017towards}) proposed a learnable metric Adem to learn from the human-annotated score of a response given its post and ground truth. However, such a metric shows very poor generalization and is not robust to easy attacks such as simple word substitution or random word shuffle \cite{sai2019re}. 
Furthermore, RUBER and its variants~\cite{tao2018ruber,ghazarian2019better} evaluate a response by directly averaging a non-learnable referenced embedding similarity score and a learnable unreferenced post-response relatedness score that is learned by applying negative sampling without human annotations. However, merely measuring input-output relatedness is not sufficient for evaluating long text generation, as the intrinsic coherence and consistency within the generated text is a critical factor. 
Additionally, some metrics which learn from human preference achieve substantial results in conditional language generation, e.g., RUSE~\cite{shimanaka2018ruse} and BLEURT~\cite{sellam2020bleurt}. RUSE trained a regression model to score a reference-candidate pair using their sentence embeddings. 
And BLEURT used multiple automatic metrics~(e.g., BLEU) as supervision signals for pretraining on synthetic data, and was fine-tuned on human judgments. However, BLEURT heavily relies on the quality of automatic metrics, but there are yet no such reliable metrics for open-ended text generation.




\section{Methodology}
\begin{figure}[!htp]
  \centering
\includegraphics[width=\linewidth]{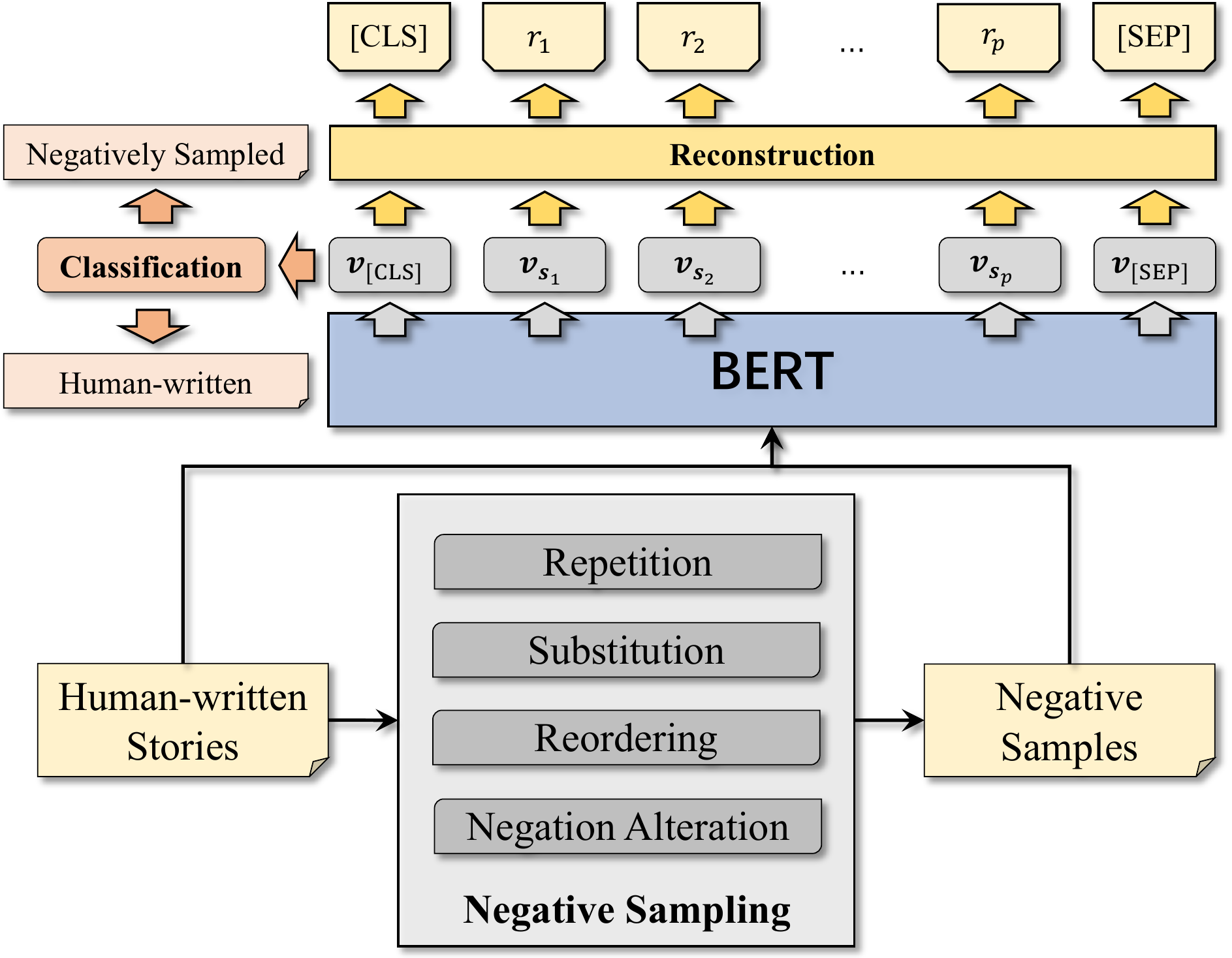}
  \caption{Overview of the \textsc{Union} metric.  \textsc{Union} is trained to distinguish the human-written stories from the negative samples  constructed by four negative sampling techniques, as well as to reconstruct the original human-written stories.}
  \label{fig:model}
\end{figure}

\textsc{Union} is expected to measure the overall quality of a generated story. In this section, we begin with common issues that can be observed in the output of NLG models. We then propose four negative sampling techniques based on the observations. Afterward, we introduce how \textsc{Union} is trained and used for story evaluation. The overall paradigm of \textsc{Union} is shown in Figure~\ref{fig:model}.

\subsection{Empirical Observations}
The key aspect of \textsc{Union} is the construction of negative samples, which provides a range of lexical, syntactic, and semantic variations to simulate the errors made by NLG models.  
Therefore, we first present our empirical observations regarding the question \textit{``What makes a story unreasonable for NLG models?''}.

We analyzed 381 unreasonable stories generated  by various NLG models like Plan\&Write~\cite{yao2018plan} and fine-tuned GPT-2~\cite{radford2019language} 
base on ROCStories~\cite{mostafazadeh2016corpus}, and summarized four major types of errors, including \textbf{repeated plots}~(repeating similar texts), \textbf{poor coherence}~(with unrelated keywords or events but a reasonable main plot), \textbf{conflicting logic}~(wrong causal or temporal relationship), and \textbf{chaotic scenes}~(difficult to understand or with multiple previous errors). To facilitate understanding of the error types, we resorted to manual annotation of all the unreasonable stories. And seven annotators were hired for each story~(see the full details in Section \ref{sec:ann}). In addition to the four error types, we also provide annotators with an option \textbf{Others}. 
We summarize the proportion of stories annotated with different error types in Table~\ref{tab:prop}\footnote{Note that these human annotations are only used in test of \textsc{Union}.}. 
\begin{table}[!ht]
\footnotesize
    \centering
    \begin{tabular}{lccccc}
    \toprule
\textbf{Type}&\textbf{Repe}&\textbf{Cohe}&\textbf{Conf}&\textbf{Chao}&\textbf{Others}\\
\midrule
\textbf{Prop~(\%)}&44.1&56.2&67.5&50.4&12.9\\
	\bottomrule
    \end{tabular}
    \caption{ Error type \textbf{Prop}ortions of 381 unreasonable stories,  including  \textbf{Repe}ated plots/poor \textbf{Cohe}rence/\textbf{Conf}licting logic/\textbf{Chao}tic scenes/\textbf{Others}.}
    \label{tab:prop}
\end{table}

We can see that the four error types are the major issues of unreasonable stories, which provides rationales of constructing negative samples for evaluating generated stories. Besides, all the Spearman correlations between every two error types are less than 0.15 (p-value > 0.01), suggesting that different error types correlate weakly with each other. Furthermore, the stories annotated with 1/2/3/4 errors constitute 23.36\%/36.48\%/34.65\%/4.46\% of the annotated stories, respectively. Most of the unreasonable stories have more than one error, which motivates us to simultaneously apply multiple sampling techniques to construct negative samples.

\subsection{Constructing Negative Samples}
We construct negative samples to cover as many aforementioned issues of unreasonable stories as possible. Since using machine-generated texts as negative samples will easily lead to poor generalization (over-fitting to specific data or model bias~\cite{garbacea2019judge}), 
we devise four negative sampling techniques to automatically construct a large number of negative samples from human-written stories as follows:

\noindent\textbf{Repetition:} Generating repetitive texts is commonly observed in many state-of-the-art NLG models~\cite{fan2018hierarchical,radford2019language}, where the models focus repeatedly on what they have recently generated,  particularly with maximum-likelihood based decoding strategies~\cite{holtzman2019curious}. 
To address the issue, we introduce lexical and sentence-level repetition to construct negative samples 
using two policies---we either repeat an N-gram~(N=1,2,3,4) in a random sentence, or randomly select a sentence to repeat and remove the following sentence to keep the sentence number unchanged. 

\noindent\textbf{Substitution:} 
The coherence of a story is mainly embodied through the relationship between keywords in the context~\cite{Elizabeth2018Neural, guan2020knowledge}. Therefore, we create incoherent samples by random keywords and sentence substitution, respectively at word level and sentence level. For word-level substitution, we replace random 15\% keywords in a story with their corresponding antonyms~(e.g., replace ``deny'' with ``confirm''), 
otherwise with another random keyword sampled from all the keywords of the same part-of-speech~(POS), according to the mention frequency. We use the commonsense knowledge base ConceptNet ~\cite{speer2012representing}\footnote{\url{http://www.conceptnet.io/}} for keyword recognition and antonym query. ConceptNet consists of commonsense triples like \texttt{(h, r, t)}, meaning that the head concept \texttt{h} has a relation \texttt{r} with the tail concept \texttt{t}, e.g., \texttt{(evaluation, IsA, judgment)}. We regard those words which are heads or tails in ConceptNet as keywords. And given an keyword, we look up those keywords as its antonyms with which have negated relations, 
including \texttt{Antonym, NotDesires, NotCapableOf}, and \texttt{NotHasProperty}. If no antonym is found for a keyword, we perform replacement with a random keyword of the same POS. And we adopt NLTK\footnote{\url{http://nltk.org/}} for POS tagging. 

For sentence-level substitution, we randomly replace a sentence in a story with another one sampled from the rest of stories in the dataset.

\noindent\textbf{Reordering:} Conflicting logic usually results from wrong causal relationship and temporal dependency in the context. Therefore, we randomly reorder the sentences in a story to create negative stories with conflicting plot.

\noindent\textbf{Negation Alteration:} Negation words such as ``not'' are crucial for language generation tasks because they may flip the semantics of a sentence, which is also an important cause of conflicting logic. We perform negation alteration by adding or removing negation words using rules for different types of verbs\footnote{The details are shown in the supplementary material.}. 

Since there may be multiple error types in a generated story, we apply different sampling techniques simultaneously to construct a negative sample. We first sample the number ($n$) of techniques from \{1,2,3,4\} with a distribution \{50\%, 20\%, 20\%, 10\%\}. We then sample a technique without replacement from \{repetition, substitution, reordering, negation alteration\} with a distribution \{10\%, 30\%, 40\%, 20\%\} until the total number of techniques ($n$) is reached. Last, we apply the sampled techniques on a human-written story to obtain a perturbated sample.
A constructed example is shown in Table~\ref{tab:neg_sample}. 



\begin{table}[!ht]
\footnotesize
    \centering
    \begin{tabular}{p{7cm}}
    \toprule
    {\textbf{Leading Context}}\\
    
    {Ken was out jogging one morning.}\\
    \midrule
    {\textbf{Reference By Human}}\\
    {The weather was crisp and cool. Ken felt good and energetic. He decided to keep jogging longer than normal. Ken went several more miles out of his way.
} \\
\midrule
    {\textbf{Auto-Constructed Negative Sample}}\\
    {The weather was crisp and cool \textit{and cool}. Ken felt \underline{bad} and energetic. \textbf{Ken \textsc{did not go} several more miles out of his way. He decided to keep jogging longer than normal}.
} \\
	\bottomrule
    \end{tabular}
    \caption{An example of negative sample construction. The repeated bigram is in \textit{italic}, the substituted keyword is \underline{underlined}, the reordered sentences are indicated in \textbf{bold}, and the altered negation words are \textsc{capitalized}.}
    \label{tab:neg_sample}
\end{table}

\subsection{Modeling}
Let $\{\bm{s}_n, \bm{r}_n,  y_n\}_{n=1}^N$ denote the training dataset of size $N$ for training the \textsc{Union} metric, where $\bm{s}_n$ is a human-written story or an auto-constructed negative sample, $\bm{r}_n$ is the corresponding original story of $\bm{s}_n$. If $\bm{s}_n$ is a negative sample, $y_n=0$, otherwise $y_n=1$ where $\bm{s}_n$ is exactly the same as $\bm{r}_n$ in this case. $y_n\in\{0,1\}$ indicates whether $\bm{s}_n$ is written by human. 
For better story understanding, 
we leverage BERT~\cite{devlin2018bert} to obtain contextualized representations of the input. Given a story $\bm{s}_n=(s_1,s_2,\cdots,s_{p})$ of length $p$ (each $s_i$ is a word), BERT outputs a sequence of contextualized vectors:
\begin{equation}
\bm{v}_{\rm [CLS]},\bm{v}_{s_1}, \cdots, \bm{v}_{s_{p}}, \bm{v}_{\rm [SEP]} = {\rm BERT}(\bm{s}_n),
\end{equation}
where $\bm{v}_{\rm [CLS]}$ and $\bm{v}_{\rm [SEP]}$ are the representation for the special tokens $\rm [CLS]$ and $\rm [SEP]$, respectively. 
We add a task-specific linear layer on top of the $\rm [CLS]$ vector to predict the \textsc{Union} score, indicating the probability that $\bm{s}_n$ is written by human:
\begin{equation}
	{\hat y}_n={\rm sigmoid}(\textbf{W}_c\bm{v}_{\rm [CLS]}+\textbf{b}_c),
\end{equation}
where $\textbf{W}_c$ and $\textbf{b}_c$ are trainable parameters. 
We use the cross entropy loss to optimize the prediction objective as follows:
\begin{equation}
	\mathcal{L}_n^{C} = -y_n~{\rm log}~{\hat y_n}-(1-y_n)~{\rm log}~(1-\hat y_n).
\end{equation}

In addition to the main prediction task, we devise an auxiliary reconstruction task which requires to reconstruct the corresponding human-written story $\bm{r}_n$ from perturbated story $\bm{s}_n$. Therefore, we add an additional linear layer at the last layer of BERT, which takes as input the vectors 
output from the last transformer block and computes a probability distribution over the entire vocabulary through a softmax layer, formally as follows:
\begin{equation}
	P(\hat{r}_i|\bm{s}_n)={\rm softmax}(\textbf{W}_r\bm{v}_{s_i}+\textbf{b}_r),
\end{equation}
where $\hat{r}_i$ is the predicted $i$-th token, $\textbf{W}_r$ and $\textbf{b}_r$ are the parameters of the additional linear layer. Then the model is trained by minimizing the negative log-likelihood:
\begin{equation}
	\mathcal{L}_n^{R}=-\frac1{p}\sum_{i=1}^{p}{\rm log}~P(\hat{r}_i=r_i|\bm{s}_n),
\end{equation}
where $r_i$ is the $i$-th token in human-written story $\bm{r}_n$. 
The combined loss function $\mathcal L$ of the full model is computed as follows:
\begin{equation}
	\mathcal{L}=\frac1N\sum_{n=1}^N(\mathcal{L}_n^C + \lambda \mathcal{L}_n^R),
\end{equation}
where $\lambda$ is an adjustable hyperparameter.

We fine-tune all the parameters of \textsc{Union} on the training dataset, including the BERT and the two additional linear layers. In practical use, \textsc{Union} can measure the quality of a new generated sample $\bm{\hat s}$ by taking $\bm{\hat s}$ as input to predict the corresponding score $\hat y$.


\section{Experiment}

We conducted extensive experiments to evaluate \textsc{Union} on two story datasets. First, we compared \textsc{Union} against existing text generation metrics. Then, we assessed its generalization on distribution drifts, including dataset drift and quality drift. Last, we measured the effect of each negative sampling technique with ablation studies.

\subsection{Baselines}
We compared \textsc{Union} with the following three kinds of metrics as baselines:
\\
\textbf{Referenced metrics:} sentence \textit{\textbf{BLEU}} score (geometric mean of 1-gram up to 4-gram)~\cite{papineni2002bleu} to measure the lexical similarity between a candidate sample and its reference, and \textit{\textbf{MoverScore}} ~\cite{zhao2019moverscore} to measure the semantic similarity.
\\
\textbf{Unreferenced metrics:}  \textit{\textbf{Perplexity}}\footnote{We take the minus of perplexity for all the following experiments to ensure a higher value means better  quality.} computed by the GPT-2 model~\cite{radford2019language}, and a discriminative evaluator ~(\textit{\textbf{DisScore}})~\cite{kannan2017adversarial} that is trained based on BERT to distinguish generated samples from human-written stories.  
\\
\textbf{Hybrid metrics}: \textit{\textbf{RUBER-BERT}}~\cite{ghazarian2019better} which improves the original RUBER~\cite{tao2018ruber} with contextualized embeddings from BERT, and the supervised metric \textit{\textbf{BLEURT}}~\cite{sellam2020bleurt} that is fine-tuned on human judgments after pretraining on large-scale synthetic data with multiple automatic metrics as supervision signals. 

In addition, we also reported the performance of the referenced and unreferenced versions in {RUBER-BERT}, denoted as \textit{\textbf{RUBER$_r$-BERT}} and \textit{\textbf{RUBER$_u$-BERT}}, respectively. 

We set the parameters of \textsc{Union} by following the uncased base version of \citeauthor{devlin2018bert}~(\citeyear{devlin2018bert}): the transformer has 12 layers, 768 dimensional hidden states, and 12 attention heads. We used batch size 10, and learning rate 5e-5. The scale factor $\lambda$ is set to 0.1. We directly used public pretrained parameters of BERT\footnote{\url{https://github.com/google-research/bert}} or GPT-2\footnote{\url{https://github.com/openai/gpt-2}}~(base version) for all the baselines. 

\subsection{Data Preparation}\label{sec:ann}
We used two datasets for evaluation, ROCStories (\textbf{ROC} for short)~\cite{mostafazadeh2016corpus} and WritingPrompts (\textbf{WP})~\cite{fan2018hierarchical}. The ROC dataset contains 98,161 five-sentence human-written stories, 
with an average length of 49.4 words. 
To achieve better generalization performance, we followed \citeauthor{guan2020knowledge}~(\citeyear{guan2020knowledge}) to make delexilization by masking all the male/female/unknown names with placeholders [MALE]/[FEMALE]/[NEUTRAL], respectively. 

The WP dataset consists of 303,358 stories paired with writing prompts collected from an online forum. 
The average length of the prompt/story is 28.4/734.5 respectively, much longer than those in ROC. Since it is still challenging for state-of-the-art NLG models to maintain a reasonable plot through the whole story, and hard to obtain acceptable annotation agreement in manual evaluation of long stories, we retained about 200 words (with correct sentence boundary) from the start and truncated the rest in WP for subsequent experiments.

We randomly selected 90\%/5\%/5\% stories from both datasets for training/validation/test of \textsc{Union} and learnable baseline metrics, and created the evaluation set for all the metrics by generating stories based on the test sets of the datasets with state-of-the-art story generation models. The story generation models include fusion convolutional seq2seq model~\cite{fan2018hierarchical}, plan\&write~\cite{yao2018plan}, fine-tuned GPT-2 ~\cite{radford2019language}, and knowledge-enhanced GPT-2~\cite{guan2020knowledge}.

The data statistics are shown in Table~\ref{tab:data}. 
The number of negative samples for learning the metrics when necessary is the same as that of human-written stories on each dataset. Specifically, we created negative samples for DisScore by generating stories with above NLG models. For RUBER$_u$-BERT, a given leading context is appended by a randomly sampled continuation.
All the stories in the evaluation set are manually labeled. In addition, we annotated another 400 stories in ROC and 200 in WP for training BLEURT\footnote{BLEURT is first initialized with the pretrained parameters~(\url{https://github.com/google-research/bleurt}) and then fine-tuned on our annotated stories.}. 
Seven annotators were hired to judge the quality of each story with a binary score  (1 for a reasonable story, and 0 otherwise). 
Furthermore, we asked annotators to label the error type of a story if it is labeled as unreasonable, including {repeated plots}, {poor coherence}, {conflicting logic}, {chaotic scenes}, and {others}. 
We resorted to Amazon Mechanical Turk (AMT) for annotation, and the average score of the seven annotators is treated as the final score. 
We provide the full details of the instruction for annotators in the supplementary file.


\begin{table}[!ht]
\small
    \centering
    \begin{tabular}{l|l|c|c|c}
    \hline
    \textbf{Split}&\textbf{Metrics}&\textbf{ROC}&\textbf{WP}&\textbf{NS}\\
    \hline
&\textbf{Perplexity}&&&\ding{55}\\
&\textbf{DisScore}&88,344/ &272,600/&\ding{51}\\
\textbf{{Train/}}&\textbf{RUBER$_u$}&4,908&15,620&\ding{51}\\
\textbf{Validate}&\textbf{\textsc{Union}}& &&\ding{51}\\
\cline{2-5}
& \textbf{BLEURT}&360$^\dagger$/40$^\dagger$ &180$^\dagger$/20$^\dagger$&\ding{55} \\
\hline
\textbf{{Test}}&\textbf{All metrics}&400$^\dagger$&200$^\dagger$&N/A\\
\hline
    \end{tabular}
    \caption{Data statistics.  \textbf{RUBER$_u$} is short for \textbf{RUBER$_u$-BERT}. \textbf{NS} (Negative Sampling) means whether a metric requires negative samples for training/validation. 
    $^\dagger$ means the stories are generated by NLG models and manually annotated.
}
    \label{tab:data}
\end{table}

\begin{table*}[!ht]
\small
    \centering
    \begin{tabular}{clcccccc}
    \toprule
\multicolumn{2}{c}{\multirow{2}{*}{{\textbf{Metrics}}}} & \multicolumn{3}{c}{\textbf{ROC}}&\multicolumn{3}{c}{\textbf{~~~~~WP}}\\
& &$r$&$\rho$ &$\tau$ &~~~~~$r$&$\rho$ &$\tau$\\
	\midrule
\multirow{2}{*}{\textbf{Referenced}}	&\textbf{BLEU} & 0.0299$^{~~}$& 0.0320$^{~~}$& 0.0231$^{~~}$& ~~~~~0.1213$^{~~}$&0.0941$^{~~}$&0.0704$^{~~}$\\
	&\textbf{MoverScore}&0.1538$^{*}$&0.1535$^{*}$&0.1093$^{*}$&~~~~~0.1613$^{~~}$&0.1450$^{~~}$&0.1031$^{~~}$\\
	&\textbf{RUBER$_r$-BERT}&0.0448$^{~~}$&0.0517$^{~~}$&0.0380$^{~~}$&~~~~~0.1502$^{~~}$&0.1357$^{~~}$&0.0986$^{~~}$\\
		\midrule
	\multirow{5}{*}{\textbf{Unreferenced}}	&\textbf{Perplexity}&0.2464$^{*}$&0.2295$^{*}$&0.1650$^{*}$&~~~~~-0.0705$^{~~}$&-0.0479$^{~~}$&-0.0345$^{~~}$\\
	&\textbf{RUBER$_u$-BERT}&0.1477$^{*}$&0.1434$^{*}$&0.1018$^{*}$&~~~~~0.1613$^{~~}$&0.1605$^{~~}$&0.1157$^{~~}$\\
	&\textbf{DisScore}&0.0406$^{~~}$&0.0633$^{~~}$&0.0456$^{~~}$&~~~~~0.0627$^{~~}$&-0.0234$^{~~}$&-0.0180$^{~~}$\\
	&\textbf{\textsc{Union}}&\textbf{0.3687$^{*}$}&\textbf{0.4599}$^{*}$&\textbf{0.3386}$^{*}$&~~~~~\textbf{0.3663}$^{*}$&\textbf{0.4493}$^{*}$&\textbf{0.3293}$^{*}$\\
	&~~~~\textbf{-Recon}&{0.3101$^{*}$}&{0.4027}$^{*}$&{0.2927}$^{*}$&~~~~~{0.3292}$^{*}$&0.3786$^{*}$&0.2836$^{*}$\\	
	\midrule
	\multirow{2}{*}{\textbf{Hybrid}}&\textbf{RUBER-BERT}&0.1412$^{*}$&0.1395$^{*}$&0.1015$^{*}$&~~~~~0.1676$^{~~}$&0.1664$^{~~}$&0.1194$^{~~}$\\
		&\textbf{BLEURT}&0.2310$^{*}$&0.2353$^{*}$&0.1679$^{*}$&~~~~~0.2229$^{*}$&0.1602$^{~~}$&0.1180$^{~~}$\\
	\bottomrule
    \end{tabular}
    \caption{Correlation with human judgments on ROC and WP datasets. $r$/$\rho$/$\tau$ indicates the Pearson/Spearman/Kendall correlation, respectively. The best performance is highlighted in \textbf{bold}. The correlation scores marked with * indicate the result significantly correlates with human judgments (p-value<0.01).}
    \label{tab:results}
\end{table*}

\subsection{Correlation Results}
Correlation analysis has been widely used to evaluate automatic metrics for language generation~\cite{tao2018ruber,sellam2020bleurt}.
We employed \textsc{Union} and other metrics to score the collected samples, and then calculated the Pearson~($r$), Spearman~($\rho$) and Kendall~($\tau$) correlation coefficients between model evaluation and human judgments.  Pearson's $r$ estimates linear correlation while Spearman's $\rho$ and Kendall's $\tau$ estimate monotonic correlation, and $\tau$ is usually more insensitive to abnormal values than $\rho$. We used the standard statistical package \texttt{stats} in \texttt{SciPy}\footnote{\url{https://docs.scipy.org/doc/scipy/reference/stats.html}} for correlation calculation and significance test.

As summarized in Table \ref{tab:results}, the referenced metrics correlate worse with human judgments, particularly for BLEU which is based on lexical similarity. Measuring the semantic similarity instead (MoverScore, RUBER$_r$-BERT) can improve the correlation but is still limited, indicating that referenced metrics are not competitive for evaluating open-ended language generation. 
Perplexity is ineffective on WP  because the generated stories in the dataset are much longer and hence suffer from more serious repetition errors than those in ROC, which easily results in low perplexity (i.e., high minus perplexity)~\cite{holtzman2019curious} but poor human judgment scores. Furthermore, \textsc{Union} outperforms other baselines including the supervised metric BLEURT by a large margin, which also demonstrates the advantage of unreferenced metrics. 
Besides, removing the reconstruction training objective~(-Recon) leads to remarkably worse correlation, indicating that the auxiliary task further improves the performance of \textsc{Union}. 

\subsection{Generalization to Dataset and Quality Drift}

It is extremely important for learnable metrics to deal with dataset drift and quality drift~\cite{sellam2020bleurt}. Specifically, a generalizable metric is expected to reliably evaluate outputs from different datasets even without re-training. Moreover, since the quality of generated samples can vary significantly across NLG models, a reliable metric should be able to evaluate samples of different quality levels. Therefore, we conducted experiments to assess the generalization ability of \textsc{Union} in this section.

\begin{table}[!h]
    \centering
    \small
    \begin{tabular}{lccc}
    \toprule
 \textbf{Metrics}
 &   $r$&$\rho$ &$\tau$\\
	\midrule\midrule
\multicolumn{4}{c}{\textbf{Training}: WP~~ \textbf{Test}: ROC}\\
\midrule
 \textbf{Perplexity}
 &-0.0015 &0.0149$^{~~}$&0.0101$^{~~}$\\
 \textbf{RUBER$_u$-BERT} &-0.0099 &-0.0162 &-0.0110 \\
\textbf{BLEURT} & 0.1326$^*$& 0.1137$^*$&0.0828$^*$ \\
\textbf{\textsc{Union}} &\textbf{0.1986}$^*$ &\textbf{0.2501}$^*$ & \textbf{0.1755}$^*$\\
\textbf{~~~~-Recon}&0.1704$^*$ & 0.2158$^*$&0.1523$^*$\\
\midrule\midrule
\multicolumn{4}{c}{\textbf{Training}: ROC~~ \textbf{Test}: WP}\\
\midrule
\textbf{Perplexity}
& 0.0366$^{~~}$&0.0198$^{~~}$&0.0150$^{~~}$\\
\textbf{RUBER$_u$-BERT} &0.1392$^{~~}$ &0.1276$^{~~}$ &0.0912$^{~~}$ \\
\textbf{BLEURT} &0.1560$^{~~}$ & 0.1305$^{~~}$& 0.0941$^{~~}$\\
\textbf{\textsc{Union}} &\textbf{0.2872}$^*$&\textbf{0.2935}$^*$&\textbf{0.2142}$^*$\\
\textbf{~~~~-Recon}& 0.2397$^*$ & 0.2712$^*$& 0.1971$^*$\\
	\bottomrule
    \end{tabular}
    \caption{Correlation results in the dataset drift setting where the metrics are trained on one dataset and then used for the other one.}
    \label{tab:dataset_transfer}
\end{table}
\begin{figure*}[t]
\centering
\subfigure{\includegraphics[width=0.5\linewidth]{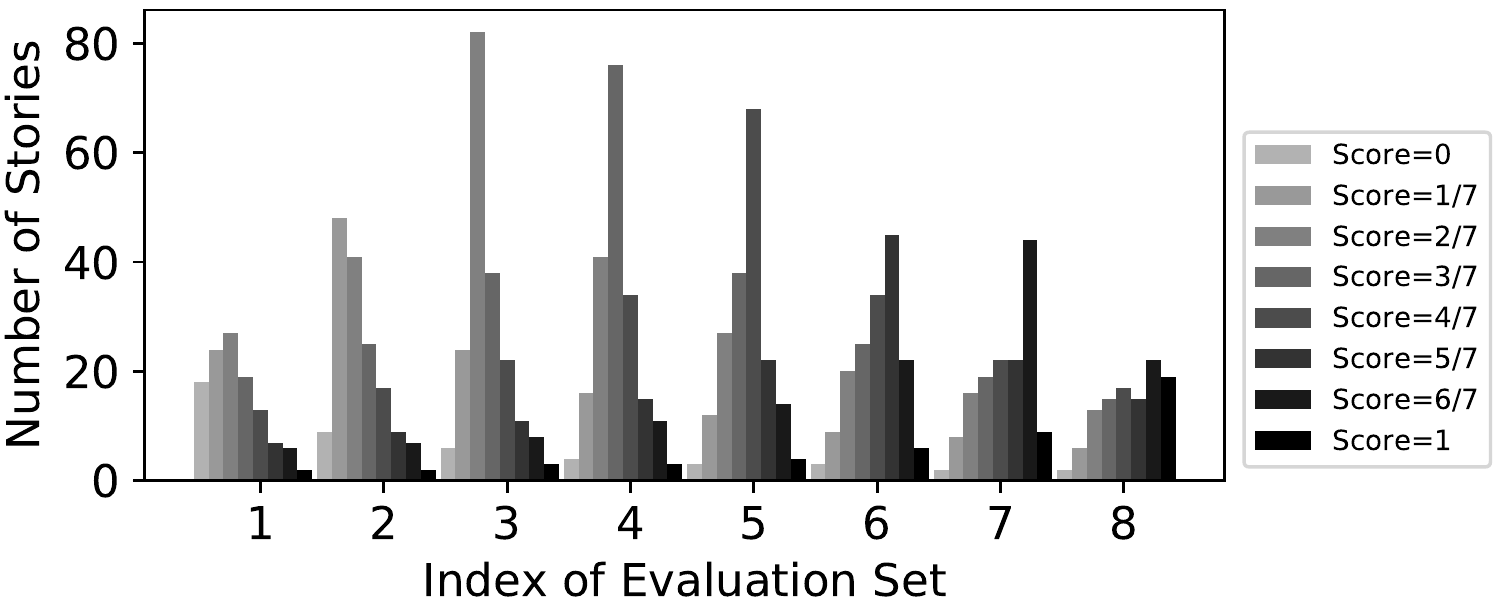}}\subfigure{\includegraphics[width=0.5\linewidth]{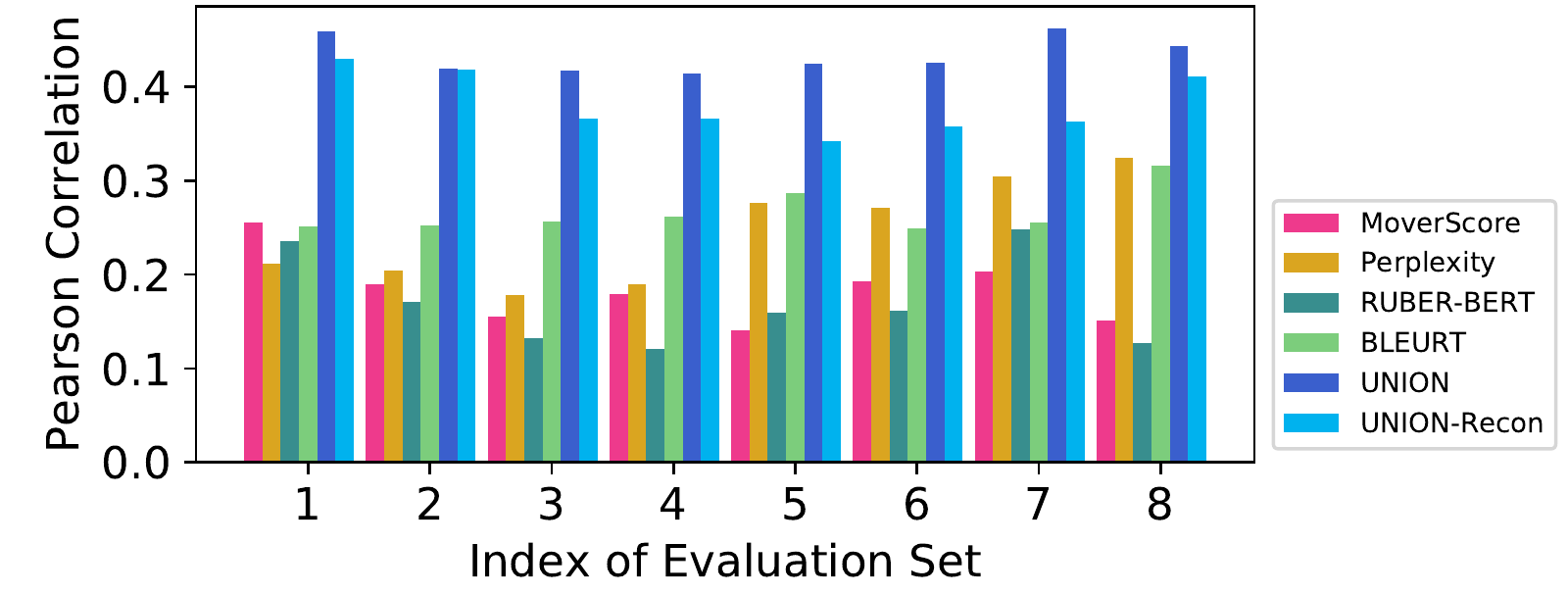}}
\vspace{-7mm}
\caption{Generalization over different biased test sets. Left: distribution of stories of different annotation scores in different test sets. Right: the Pearson correlation of different metrics with human judgments on different test sets, where {\scriptsize\textsf{UNION-Recon}} denotes \textsc{Union} without the reconstruction task.}
\label{fig:quality}
\end{figure*}

\begin{table*}[!ht]
\small
    \centering
    \begin{tabular}{lccccc}
    \toprule
    
\multirow{2}{*}{\textbf{Evaluation Set}}&\multirow{2}{*}{\textbf{All Samples~(400)}}&\multicolumn{4}{c}{\textbf{Reasonable Samples~(19) + Unreasonable Samples with}}\\ 
&&\textbf{Repe~(24)} & \textbf{Cohe~(38)} & \textbf{Conf~(61)} & \textbf{Chao~(23)}\\ 
    \midrule
	\textbf{\textsc{Union}}&0.3687& 0.6943&0.5144&0.4571&0.6744\\
	\textbf{~~~~-Repetition}&0.3167~({{\textbf{\scriptsize $\downarrow$14\%}}})& 0.4743~(\textbf{{\scriptsize $\downarrow$32\%}})&0.5308~~({{\scriptsize $\uparrow$3\%}})&0.4316~~({{\scriptsize $\downarrow$6\%}})&0.6561~~({{\scriptsize $\downarrow$3\%}})\\
	\textbf{~~~~-Substitution}&0.3118~(\textbf{{{\scriptsize $\downarrow$15\%}}})& 0.7034~~({{\scriptsize $\uparrow$1\%}})&0.4185~(\textbf{{{\scriptsize $\downarrow$19\%}}})&0.4468~~({{\scriptsize $\downarrow$2\%}})&0.5850~({\textbf{{\scriptsize $\downarrow$13\%}}})\\
	\textbf{~~~~-Reordering}&0.2302~({\textbf{{\scriptsize $\downarrow$38\%}}})& 0.6546~~({{\scriptsize $\downarrow$6\%}})&0.5077~~({{\scriptsize $\downarrow$1\%}})&0.3507~(\textbf{{{\scriptsize $\downarrow$23\%}}})&0.5393~({\textbf{{\scriptsize $\downarrow$20\%}}})\\	
	\textbf{~~~~-Negation Alteration} &0.3304~({\textbf{{\scriptsize $\downarrow$10\%}}})&0.6665~~({{\scriptsize $\downarrow$4\%}})&0.4987~~({{\scriptsize $\downarrow$3\%}})&0.3946~(\textbf{{{\scriptsize $\downarrow$14\%}}})&0.5176~(\textbf{{{\scriptsize $\downarrow$23\%}}})\\
	\bottomrule
    \end{tabular}
    \caption{ Pearson correlation with different negative sampling techniques. The numbers in parentheses denote the number  of stories. The error types include \textbf{Repe}ated plots, poor \textbf{Cohe}rence, \textbf{Conf}licting logic, and \textbf{Chao}tic scenes. The proportions in parentheses indicate the relative change with respect to \textsc{Union} (the first row).}
    \label{tab:ablation}
\end{table*}

To assess the generalization to dataset drift, we first trained the learnable metrics on ROC and then directly used them to evaluate generated stories from WP, and vise versa. Table~\ref{tab:dataset_transfer} shows the Pearson correlation with human judgments in this setting. 
Compared with the results in Table \ref{tab:results}, all the metrics trained on one dataset have remarkable drops in correlation when they are used for the other dataset because the two datasets are significantly different in length and topic. Nevertheless, \textsc{Union} performs more robustly than other metrics, with much better correlation with human judgments. Moreover, our method of constructing negative examples is generalizable to the two datasets.

To assess the generalization of \textsc{Union} to quality drift, we created biased test sets from ROC by sampling stories of different quality levels with different probabilities.
Specifically, the annotation score of each story ranges from 0 to 1 (i.e., $0, \frac17,\frac27,\cdots,1$) since there are seven annotators for each sample. We then created 8 biased sets, indexed from 1 to 8 with variable $I$. For the $I^{th}$ set, we sampled the stories whose annotation score is $\frac k7$ with a probability of $\frac{1}{|I-k|+1}$ where $k\in\{0,1,\cdots,7\}$. 
In this way, the 8 sets have different distributions of stories with different qualities\footnote{
We assume that the annotation score $\frac k7$ approximates the quality level.}, as shown in Figure \ref{fig:quality} (left). 

We then computed the Pearson correlation of different metrics with human judgments on the 8 sets. 
Results in Figure \ref{fig:quality}~(right) show that: \textbf{I.} \textsc{Union} has higher correlation than other metrics on all the biased sets.
\textbf{II.} \textsc{Union} is more reliable and robust than other metrics, with much less variance. For instance, MoverScore performs much better on Set \#1 (with more low-quality stories) than on Set \#8 (with more high-quality stories). Interestingly,
 Perplexity performs much better on high-quality sets than on low-quality ones, because high-quality stories are closer to human-written stories from which a language model learns. 
 \textbf{III.} The ablated \textsc{Union} without the reconstruction objective has lower correlation and larger variance, indicating that the auxiliary task can improve the discriminative and generalization ability.


\subsection{Ablation Studies}
To understand the effect of each negative sampling technique, we conducted ablation tests on ROC dataset. Each time we ablated one technique of constructing negative samples, re-trained \textsc{Union} on the constructed data, and evaluated it on five evaluation sets: all 400 samples, and four other sets where each contains 19 reasonable samples and other unreasonable samples of some error type. The error type of a story is decided if at least three of seven annotators annotate the same error type. 

Table \ref{tab:ablation} shows the Pearson correlation results. \textsc{Union} is remarkably better than its ablated version on the all-sample set, indicating the necessity of the four techniques for constructing negative samples. Reordering seems to be the most important technique, which agrees with our observation that conflicting logic is the major issue in existing story generation models. Furthermore, as expected, the correlation drops remarkably on the evaluation set of some error type if without the corresponding negative sampling technique. Interestingly, it is easier for \textsc{Union} to evaluate repetitive/chaotic stories, which seem to be easier cases in story generation.



\section{Conclusion}
We present \textsc{Union}, an unreferenced metric for evaluating open-ended story generation. \textsc{Union} is trained to distinguish human-written stories from auto-constructed negative samples and to recover the perturbation in negative samples.
Extensive experiments show that \textsc{Union} outperforms state-of-the-art metrics in terms of correlation with human judgments on two story datasets, and is more robust to dataset drift and quality drift. Results also show the effectiveness of the proposed four negative sampling techniques.
As future work, we will explore the similar idea of designing unreferenced metrics for dialog generation.


\section*{Acknowledgments}

This work was jointly supported by the NSFC projects (Key project with No. 61936010 and regular project with No. 61876096), and the Guoqiang Institute of Tsinghua University，with Grant No. 2019GQG1.
We thank THUNUS NExT Joint-Lab for the support.

\bibliography{emnlp2020}

\begin{thebibliography}{29}
\expandafter\ifx\csname natexlab\endcsname\relax\def\natexlab#1{#1}\fi

\bibitem[{Bahdanau et~al.(2015)Bahdanau, Cho, and Bengio}]{bahdanau2015neural}
Dzmitry Bahdanau, Kyunghyun Cho, and Yoshua Bengio. 2015.
\newblock Neural machine translation by jointly learning to align and
  translate.
\newblock In \emph{3rd International Conference on Learning Representations,
  ICLR 2015}.

\bibitem[{Clark et~al.(2018)Clark, Ji, and Smith}]{Elizabeth2018Neural}
Elizabeth Clark, Yangfeng Ji, and Noah~A. Smith. 2018.
\newblock Neural text generation in stories using entity representations as
  context.
\newblock In \emph{NAACL}, pages 1631--1640.

\bibitem[{Devlin et~al.(2019)Devlin, Chang, Lee, and
  Toutanova}]{devlin2018bert}
Jacob Devlin, Ming-Wei Chang, Kenton Lee, and Kristina Toutanova. 2019.
\newblock Bert: Pre-training of deep bidirectional transformers for language
  understanding.
\newblock In \emph{Proceedings of the 2019 Conference of the North American
  Chapter of the Association for Computational Linguistics: Human Language
  Technologies, Volume 1 (Long and Short Papers)}, pages 4171--4186.

\bibitem[{Fan et~al.(2018)Fan, Lewis, and Dauphin}]{fan2018hierarchical}
Angela Fan, Mike Lewis, and Yann Dauphin. 2018.
\newblock Hierarchical neural story generation.
\newblock In \emph{Proceedings of the 56th Annual Meeting of the Association
  for Computational Linguistics (Volume 1: Long Papers)}, pages 889--898.

\bibitem[{Garbacea et~al.(2019)Garbacea, Carton, Yan, and
  Mei}]{garbacea2019judge}
Cristina Garbacea, Samuel Carton, Shiyan Yan, and Qiaozhu Mei. 2019.
\newblock Judge the judges: A large-scale evaluation study of neural language
  models for online review generation.
\newblock In \emph{Proceedings of the 2019 Conference on Empirical Methods in
  Natural Language Processing and the 9th International Joint Conference on
  Natural Language Processing (EMNLP-IJCNLP)}, pages 3959--3972.

\bibitem[{Ghazarian et~al.(2019)Ghazarian, Wei, Galstyan, and
  Peng}]{ghazarian2019better}
Sarik Ghazarian, Johnny Wei, Aram Galstyan, and Nanyun Peng. 2019.
\newblock Better automatic evaluation of open-domain dialogue systems with
  contextualized embeddings.
\newblock In \emph{Proceedings of the Workshop on Methods for Optimizing and
  Evaluating Neural Language Generation}, pages 82--89.

\bibitem[{Guan et~al.(2020)Guan, Huang, Zhao, Zhu, and
  Huang}]{guan2020knowledge}
Jian Guan, Fei Huang, Zhihao Zhao, Xiaoyan Zhu, and Minlie Huang. 2020.
\newblock A knowledge-enhanced pretraining model for commonsense story
  generation.
\newblock \emph{Transactions of the Association for Computational Linguistics},
  8:93--108.

\bibitem[{Hashimoto et~al.(2019)Hashimoto, Zhang, and
  Liang}]{hashimoto2019unifying}
Tatsunori Hashimoto, Hugh Zhang, and Percy Liang. 2019.
\newblock Unifying human and statistical evaluation for natural language
  generation.
\newblock In \emph{Proceedings of the 2019 Conference of the North American
  Chapter of the Association for Computational Linguistics: Human Language
  Technologies, Volume 1 (Long and Short Papers)}, pages 1689--1701.

\bibitem[{Holtzman et~al.(2020)Holtzman, Buys, Du, Forbes, and
  Choi}]{holtzman2019curious}
Ari Holtzman, Jan Buys, Li~Du, Maxwell Forbes, and Yejin Choi. 2020.
\newblock \href {https://openreview.net/forum?id=rygGQyrFvH} {The curious case
  of neural text degeneration}.
\newblock In \emph{International Conference on Learning Representations}.

\bibitem[{Kannan and Vinyals(2017)}]{kannan2017adversarial}
Anjuli Kannan and Oriol Vinyals. 2017.
\newblock Adversarial evaluation of dialogue models.
\newblock \emph{arXiv preprint arXiv:1701.08198}.

\bibitem[{Lin(2004)}]{lin-2004-rouge}
Chin-Yew Lin. 2004.
\newblock \href {https://www.aclweb.org/anthology/W04-1013} {{ROUGE}: A package
  for automatic evaluation of summaries}.
\newblock In \emph{Text Summarization Branches Out}, pages 74--81, Barcelona,
  Spain. Association for Computational Linguistics.

\bibitem[{Liu et~al.(2016)Liu, Lowe, Serban, Noseworthy, Charlin, and
  Pineau}]{liu2016not}
Chia-Wei Liu, Ryan Lowe, Iulian~Vlad Serban, Mike Noseworthy, Laurent Charlin,
  and Joelle Pineau. 2016.
\newblock How not to evaluate your dialogue system: An empirical study of
  unsupervised evaluation metrics for dialogue response generation.
\newblock In \emph{Proceedings of the 2016 Conference on Empirical Methods in
  Natural Language Processing}, pages 2122--2132.

\bibitem[{Lowe et~al.(2017)Lowe, Noseworthy, Serban, Angelard-Gontier, Bengio,
  and Pineau}]{lowe2017towards}
Ryan Lowe, Michael Noseworthy, Iulian~Vlad Serban, Nicolas Angelard-Gontier,
  Yoshua Bengio, and Joelle Pineau. 2017.
\newblock Towards an automatic turing test: Learning to evaluate dialogue
  responses.
\newblock In \emph{Proceedings of the 55th Annual Meeting of the Association
  for Computational Linguistics (Volume 1: Long Papers)}, pages 1116--1126.

\bibitem[{Mostafazadeh et~al.(2016)Mostafazadeh, Chambers, He, Parikh, Batra,
  Vanderwende, Kohli, and Allen}]{mostafazadeh2016corpus}
Nasrin Mostafazadeh, Nathanael Chambers, Xiaodong He, Devi Parikh, Dhruv Batra,
  Lucy Vanderwende, Pushmeet Kohli, and James Allen. 2016.
\newblock A corpus and cloze evaluation for deeper understanding of commonsense
  stories.
\newblock In \emph{Proceedings of NAACL-HLT}, pages 839--849.

\bibitem[{Papineni et~al.(2002)Papineni, Roukos, Ward, and
  Zhu}]{papineni2002bleu}
Kishore Papineni, Salim Roukos, Todd Ward, and Wei-Jing Zhu. 2002.
\newblock Bleu: a method for automatic evaluation of machine translation.
\newblock In \emph{Proceedings of the 40th annual meeting on association for
  computational linguistics}, pages 311--318. Association for Computational
  Linguistics.

\bibitem[{Radford et~al.(2019)Radford, Wu, Child, Luan, Amodei, and
  Sutskever}]{radford2019language}
Alec Radford, Jeffrey Wu, Rewon Child, David Luan, Dario Amodei, and Ilya
  Sutskever. 2019.
\newblock Language models are unsupervised multitask learners.
\newblock \emph{OpenAI Blog}, 1(8).

\bibitem[{Sai et~al.(2019)Sai, Gupta, Khapra, and Srinivasan}]{sai2019re}
Ananya~B Sai, Mithun~Das Gupta, Mitesh~M Khapra, and Mukundhan Srinivasan.
  2019.
\newblock Re-evaluating adem: A deeper look at scoring dialogue responses.
\newblock In \emph{Proceedings of the AAAI Conference on Artificial
  Intelligence}, volume~33, pages 6220--6227.

\bibitem[{Sellam et~al.(2020)Sellam, Das, and Parikh}]{sellam2020bleurt}
Thibault Sellam, Dipanjan Das, and Ankur Parikh. 2020.
\newblock \href {https://doi.org/10.18653/v1/2020.acl-main.704} {{BLEURT}:
  Learning robust metrics for text generation}.
\newblock In \emph{Proceedings of the 58th Annual Meeting of the Association
  for Computational Linguistics}, pages 7881--7892, Online. Association for
  Computational Linguistics.

\bibitem[{Semeniuta et~al.(2019)Semeniuta, Severyn, and
  Gelly}]{semeniuta2018accurate}
Stanislau Semeniuta, Aliaksei Severyn, and Sylvain Gelly. 2019.
\newblock \href {https://openreview.net/forum?id=rJMcdsA5FX} {On accurate
  evaluation of {GAN}s for language generation}.

\bibitem[{Shimanaka et~al.(2018)Shimanaka, Kajiwara, and
  Komachi}]{shimanaka2018ruse}
Hiroki Shimanaka, Tomoyuki Kajiwara, and Mamoru Komachi. 2018.
\newblock Ruse: Regressor using sentence embeddings for automatic machine
  translation evaluation.
\newblock In \emph{Proceedings of the Third Conference on Machine Translation:
  Shared Task Papers}, pages 751--758.

\bibitem[{Speer and Havasi(2012)}]{speer2012representing}
Robert Speer and Catherine Havasi. 2012.
\newblock Representing general relational knowledge in conceptnet 5.
\newblock In \emph{LREC}, pages 3679--3686.

\bibitem[{Sutskever et~al.(2014)Sutskever, Vinyals, and
  Le}]{sutskever2014sequence}
Ilya Sutskever, Oriol Vinyals, and Quoc~V Le. 2014.
\newblock Sequence to sequence learning with neural networks.
\newblock In \emph{Advances in neural information processing systems}, pages
  3104--3112.

\bibitem[{Tao et~al.(2018)Tao, Mou, Zhao, and Yan}]{tao2018ruber}
Chongyang Tao, Lili Mou, Dongyan Zhao, and Rui Yan. 2018.
\newblock Ruber: An unsupervised method for automatic evaluation of open-domain
  dialog systems.
\newblock In \emph{Thirty-Second AAAI Conference on Artificial Intelligence}.

\bibitem[{Vaswani et~al.(2017)Vaswani, Shazeer, Parmar, Uszkoreit, Jones,
  Gomez, Kaiser, and Polosukhin}]{vaswani2017attention}
Ashish Vaswani, Noam Shazeer, Niki Parmar, Jakob Uszkoreit, Llion Jones,
  Aidan~N Gomez, {\L}ukasz Kaiser, and Illia Polosukhin. 2017.
\newblock Attention is all you need.
\newblock In \emph{Advances in neural information processing systems}, pages
  5998--6008.

\bibitem[{Yao et~al.(2019)Yao, Peng, Weischedel, Knight, Zhao, and
  Yan}]{yao2018plan}
Lili Yao, Nanyun Peng, Ralph Weischedel, Kevin Knight, Dongyan Zhao, and Rui
  Yan. 2019.
\newblock Plan-and-write: Towards better automatic storytelling.
\newblock In \emph{Proceedings of the AAAI Conference on Artificial
  Intelligence}, volume~33, pages 7378--7385.

\bibitem[{Zhang* et~al.(2020)Zhang*, Kishore*, Wu*, Weinberger, and
  Artzi}]{zhang2019bertscore}
Tianyi Zhang*, Varsha Kishore*, Felix Wu*, Kilian~Q. Weinberger, and Yoav
  Artzi. 2020.
\newblock \href {https://openreview.net/forum?id=SkeHuCVFDr} {Bertscore:
  Evaluating text generation with bert}.
\newblock In \emph{International Conference on Learning Representations}.

\bibitem[{Zhao et~al.(2017)Zhao, Zhao, and Eskenazi}]{zhao2017learning}
Tiancheng Zhao, Ran Zhao, and Maxine Eskenazi. 2017.
\newblock Learning discourse-level diversity for neural dialog models using
  conditional variational autoencoders.
\newblock In \emph{Proceedings of the 55th Annual Meeting of the Association
  for Computational Linguistics (Volume 1: Long Papers)}, pages 654--664.

\bibitem[{Zhao et~al.(2019)Zhao, Peyrard, Liu, Gao, Meyer, and
  Eger}]{zhao2019moverscore}
Wei Zhao, Maxime Peyrard, Fei Liu, Yang Gao, Christian~M Meyer, and Steffen
  Eger. 2019.
\newblock Moverscore: Text generation evaluating with contextualized embeddings
  and earth mover distance.
\newblock In \emph{Proceedings of the 2019 Conference on Empirical Methods in
  Natural Language Processing and the 9th International Joint Conference on
  Natural Language Processing (EMNLP-IJCNLP)}, pages 563--578.

\bibitem[{Zhou and Xu(2020)}]{zhou2020learning}
Wangchunshu Zhou and Ke~Xu. 2020.
\newblock Learning to compare for better training and evaluation of open domain
  natural language generation models.
\newblock In \emph{AAAI}, pages 9717--9724.

\end{thebibliography}
\bibliographystyle{acl_natbib}
\appendix

\section{Negation Alteration Rules}
We designed elaborate rules for negation alteration. The transformation rule from affirmative sentences to negative is shown in Table~\ref{tab:na}. In reverse, from negative sentences to affirmative, we removed the negation words~(``not'' or ``n't'') and altered the corresponding forms of the verbs.

Although there are other words which have negative meanings~(e.g., ``nobody''), they can be altered by another negative sampling technique: substitution with antonyms (e.g., replace ``nobody'' with ``somebody''). Therefore, we did not process these words while performing negation alteration.

\section{Annotation Instruction} 
\begin{figure*}[!htp]
  \centering
\includegraphics[width=\linewidth]{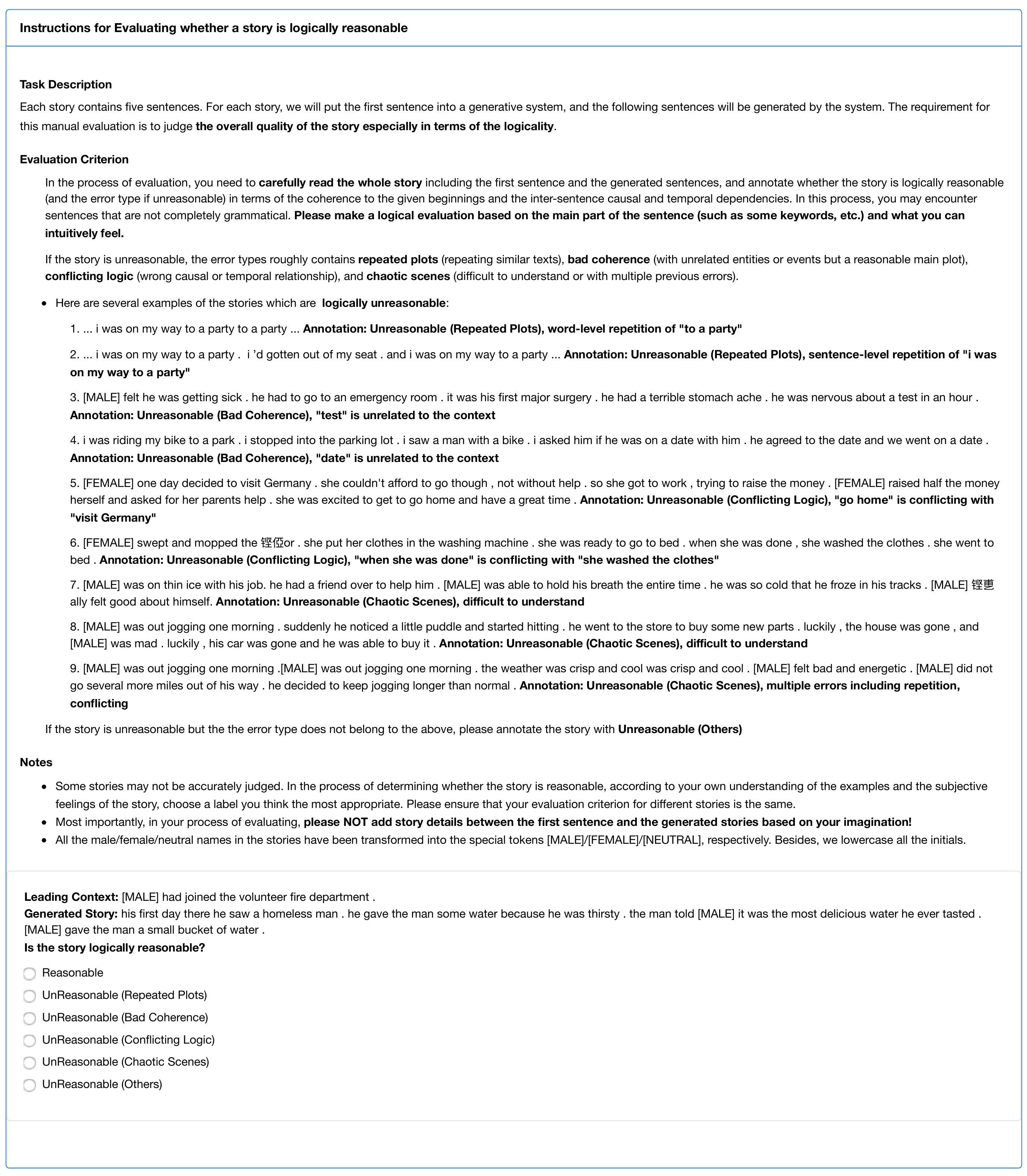}
  \caption{A screenshot of the annotation on AMT for manual evaluation.}
  \label{fig:amt}
\end{figure*}

We show a screenshot of the annotation on AMT for a generated story given a leading context from ROCStroies in Figure~\ref{fig:amt}. The annotation instruction for WritingPrompts is similar. 

\section{Annotation Results}
\begin{figure}[!htp]
  \centering
\includegraphics[width=\linewidth]{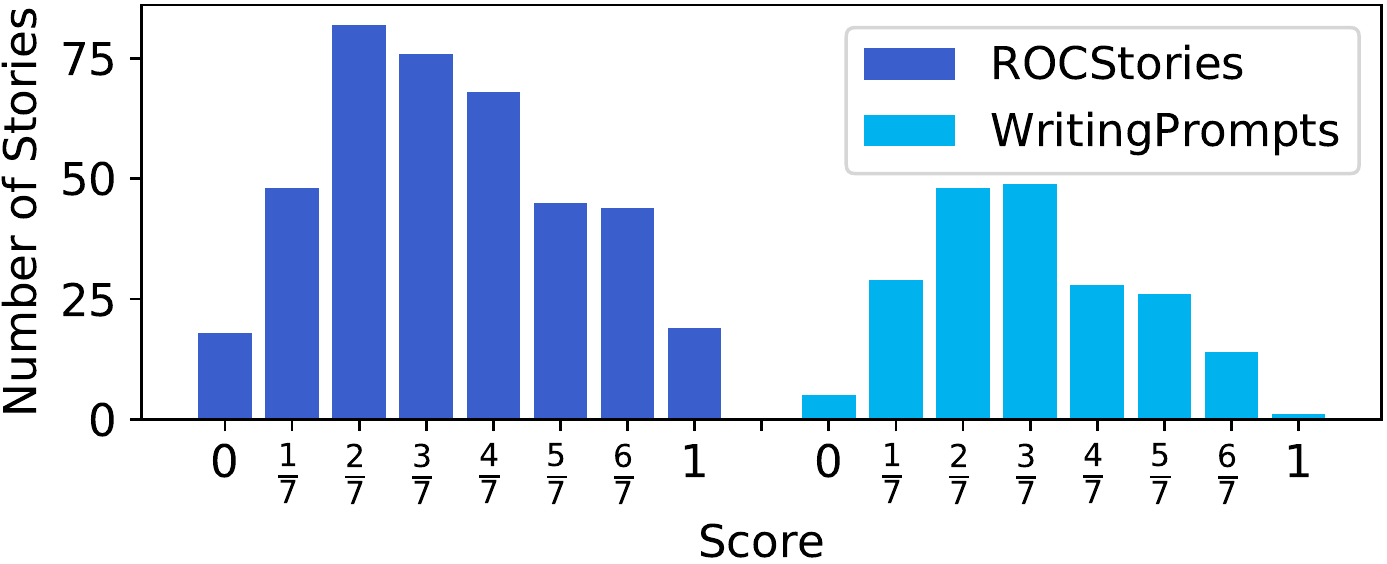}
  \caption{Number distribution of annotated stories with different human annotation scores. The total number for ROCStories/WritingPrompts is 400/200, respectively.}
  \label{fig:num}
\end{figure}

We averaged the scores of seven annotators as the final score for each story. Therefore, the annotation score ranges from $0$ to $1$ (i.e., $0, \frac17, \frac27, \cdots, 1$). The number distribution of stories with different scores is shown in Figure~\ref{fig:num}. Besides, we show 8 typical samples, one for each score in Table \ref{tab:case_score}.

\section{Reconstruction performance}
Besides the prediction objective, We also trained \textsc{Union} with an auxiliary reconstruction task, which recovers the perturbation from a negative sample. During testing, we compute the Spearman correlation between human judgments and \textsc{Union}’s editing behavior. We measure the editing behavior by labeling 1 if \textsc{Union} edits the input story, otherwise 0 if \textsc{Union} just outputs the same story. The correlation is 0.1990 (p-value<0.01) on the whole test set of ROCStories, and is 0.3442/0.1652/0.1623/0.2943 on the evaluation set which contains repetitive/incoherent/conﬂicting/chaotic stories (the same setting with Table \ref{tab:ablation}, and each set is mixed with reasonable stories), respectively. Results show that it is easier to recognize the repetitive/chaotic stories, which agrees with the results in Table \ref{tab:ablation}. 

As for the editing output, although the key motivation of the reconstruction task is to provide more speciﬁc supervision signals for recognizing errors, \textsc{Union} can generate meaningful editing results from unreasonable stories. We observe that \textsc{Union} can correct lexical errors. For example, given the story "[FEMALE] worked real hard", \textsc{Union} changed "real" to "really". However, since \textsc{Union} adopted a non-autoregressive generative framework
, it is diﬃcult to generate a grammatical story if the input has sentence-level errors. But \textsc{Union} can still accurately recognize the errors. For example, given the repetitive story "we had a great time. we had a great time.", it generated "we had a great time. we . .". We plan to improve the design by aligning the input and output tokens and then auto-tagging with editing operations during training with the reconstruction task in the future.

\section{Case Study}
We present several samples based on ROCStories and the corresponding judgments of different metrics in Table~\ref{tab:case}. We can see that it is difficult for baseline metrics to recognize the possible issues in stories,  
which rate the typical unreasonable stories~(S2-S5) even higher than the reasonable one~(S1). In comparison, \textsc{Union} judges the quality of a story more accurately regardless of whether it is similar to the reference, suggesting that \textsc{Union} can alleviate the one-to-many issue more effectively than referenced metrics~(e.g., MoverScore). For instance, although S2 maintains a reluctantly reasonable plot through the story except for a repetitive sentence, annotators still 
give it zero because there is no such repetition error in human-written stories. And \textsc{Union} successfully recognizes the issue thanks to the proposed negative sampling techniques which mimic the errors commonly observed in NLG models. Therefore, \textsc{Union} 
is more reliable for evaluating open-ended story generation.

\begin{table*}[!tp]
    \centering
    \begin{tabular}{llll}
    \toprule
    \textbf{Verb Types}&\textbf{Verb Examples}&\textbf{Rules}&\textbf{Sentence Examples}\\
    \midrule
    \textbf{be}&am, was, been, ...&\multirow{2}{*}{$\sim$ + not}&Failure \textsc{was not} an option.\\
    \textbf{Modal verbs}&would, will,  shall, ...&&I \textsc{can not} walk well.\\
    \midrule
    \textbf{Base}&go&do not + v.&I \textsc{do not go} through the park.\\
    \textbf{3rd singular present}&goes&does not + v.&He \textsc{does not go} through the park.\\
    \textbf{Simple past}&went&did not + v.&He \textsc{did not go} through the park.\\
    \midrule
    \textbf{Past participle}&gone&\multirow{2}{*}{not + $\sim$}&His insurance rate had \textsc{not gone} up.\\
    \textbf{Gerund}&going&&She ended up \textsc{not going} elsewhere.\\
	\bottomrule
    \end{tabular}
    \caption{Transformation rules from affirmative sentences to negative by adding negation words for different types of verbs. \textbf{$\sim$} stands for the current verb while \textbf{v.} for the base form of the verb. The \textsc{capitalized} words in sentence examples indicate the altered results. The negation word ``not'' can be randomly replaced with the short form ``n't''.}
    \label{tab:na}
  \end{table*}

\section{Error Analysis}
Although \textsc{Union} outperforms the
state-of-the-art metrics, it needs to be noted that the correlation with human judgments is still at a low level. As shown in Table~\ref{tab:error}, we present some typical cases where generated stories are misjudged by \textsc{Union}. Firstly, although the proposed perturbation techniques have provided many lexical and syntactic variations, it is still hard to recognize some errors such as semantic repetition and emotionally conflicting (S6-S9). Secondly, we observed that \textsc{Union} may not predict some reasonable stories (e.g., S10). This could be because some perturbated stories are still reasonable. For example, exchanging the order of two sentences without speciﬁc temporal relation (e.g., ``he had to go through a lot of training'' and ``he took a first responder's course'') does not break the story’s coherence. Training with such noisy samples may make \textsc{Union} misjudge some reasonable stories. Therefore, as future work, it is worth to explore more perturbation techniques for negative sample construction to reduce noise and cover more error types that \textsc{Union} fails to recognize. Besides, it is necessary to introduce external knowledge to help judge the logic of stories.

\begin{table*}[!htp]
\small
    \centering
    \begin{tabular}{cp{280pt}cccc}    
\toprule
\textbf{Scores}&\textbf{Samples}&\textbf{Repe}&\textbf{Cohe}&\textbf{Conf}&\textbf{Chao}\\
\midrule
\textbf{\normalsize{$0$}}&\textbf{[MALE] went to work for his father's business.} He was very careful with his business. He \texttt{\textit{didn't get into trouble}} for his mistakes. His father found out and \texttt{\textit{fired him}}. He \texttt{\textit{was a bit sad}} but \texttt{\textit{never did}}.& & & \ding{51}& \\
\midrule
\textbf{\normalsize{$\frac17$}}&\textbf{[NEUTRAL] and [MALE] had been dating for a little while.} \texttt{\textit{One day, [NEUTRAL] was convinced he could get a kiss. [NEUTRAL] decided to give her a kiss.}} They agreed to \texttt{\textit{drink it}} together. they had a good time at \texttt{\textit{the big bar}}.& & \ding{51}& \ding{51}&\ding{51}\\
\midrule
\normalsize{$\frac27$}&\textbf{[FEMALE] noticed a bird's nest \texttt{\textit{by her bedroom window.}}} She decided to \texttt{\textit{climb}} the tree. She \texttt{\textit{climbed on the ladder}} and \texttt{\textit{climbed}} up to the window. She \texttt{\textit{climbed down the ladder}} and \texttt{\textit{saw her step head}}. She reached into \texttt{\textit{her pocket}} and grabbed the bird's back. & \ding{51}& \ding{51}& \ding{51}&\ding{51}\\
\midrule
\normalsize{$\frac37$}&\textbf{[MALE]'s narcissist girlfriend only cared about what he had to offer her.} \texttt{\textit{He was a successful businessman}} who couldn't help but feed her desire. He did his best to show her that he was the real deal. She eventually left him because \texttt{\textit{he was a failure}}. Although she left him, she never found someone else to love. & & &\ding{51} &\\
\midrule
\normalsize{$\frac47$}&\textbf{[MALE] rented \texttt{\textit{an old apartment.}}} He was very bored. He \texttt{\textit{was watching}} a movie \texttt{\textit{while he watched}} it. [MALE] asked the friend to watch it. [MALE] happily watched it. & &\ding{51}&\ding{51}&\ding{51}\\
\midrule
\normalsize{$\frac57$}& \textbf{One day [FEMALE] needed to \texttt{\textit{leave the airport.}}} She was \texttt{\textit{waiting for her husband}} to get out of work. He had a bad day at work. He asked her to \texttt{\textit{meet him at the airport.}} [FEMALE] met her husband and they got in a taxi .& & & \ding{51}&\\ 
\midrule
\normalsize{$\frac67$}&\textbf{[FEMALE] saw a \texttt{\textit{smoothie}} at the store.} She saw a \texttt{\textit{chocolate cone}}. she decided to buy it. She went and bought it. The \texttt{\textit{chocolate ice cream}} was delicious. & &\ding{51} & &\\
\midrule
\normalsize{$1$}& \textbf{[MALE] had joined the volunteer fire department.} His first day there he saw a homeless man. He gave the man some water because he was thirsty. The man told [MALE] it was the most delicious water he ever tasted. [MALE] gave the man a small bucket of water.& & & & \\
	\bottomrule
    \end{tabular}
    \caption{Story samples for different human annotation scores and the annotated error types, including \textbf{{Repe}}ated plots, poor \textbf{{Cohe}}rence, \textbf{{Conf}}licting, and \textbf{{Chao}}tic scenes. \textbf{Bold} sentences are the given leading context. \textit{\texttt{Italic}} words denote the improper entities or events.}
    \label{tab:case_score}
\end{table*}

\begin{table*}[h]
\footnotesize
    \centering
    \begin{tabular}{p{10pt}p{80pt}p{105pt}p{105pt}p{12pt}p{12pt}p{12pt}p{12pt}}
    \toprule
\textbf{ID}&\textbf{Leading Context}&\textbf{Reference}&\textbf{Generated Samples}& \underline{\textbf{H}} &  \textbf{M}&\textbf{B}&\textbf{\textsc{U}}\\
\midrule
S1
&{[MALE] had joined the volunteer fire department.} 
& He had to go through a lot of training. He took a first responder's course. [MALE] was first to respond on a scene one time. He saved a man 's life. &{His first day there he saw a homeless man. He gave the man some water because he was thirsty. The man told [MALE] it was the most delicious water he ever tasted. [MALE] gave the man a small bucket of water.}
& \underline{1.00} & 0.34 & 0.43 & 0.99\\
\midrule
S2&{We were looking for something fun to do on a Tuesday night.}&We decided to see a new movie that was out. When we got there we found out the tickets were half price on Tuesdays. We decided Tuesdays will now be our standing date night. It is such a nice, fun, cheap night that we can look forward to. &{ My wife and I were so excited. We went to the mall. \texttt{\textit{We had a great time. We had a great time.}}} \textit{\texttt{(Repe)}}& \underline{0.00} &0.44 &0.49 &0.00\\
\midrule
S3&{[NEUTRAL] had a new baby brother.}&The baby would cry all night. [NEUTRAL] wasn't able to sleep. [NEUTRAL] started to despise his brother. He asked his mom if he could move to his grandmother's. & He wanted to do something new. He was sad to see other kids and play his own. [NEUTRAL] had a great time. [NEUTRAL]'s dad decided to \texttt{\textit{go shopping}}. \texttt{\textit{(Cohe)}} & \underline{0.00} & 0.48 & 0.54 & 0.00\\
\midrule
S4&{[MALE] went to work for his father's business.}& His father was the boss. [MALE] was lazy at work. Everyone was scared to tell his father. [MALE] continued to do a bad job. &  He was very careful with his business. He \texttt{\textit{didn't get into trouble}} for his mistakes. His father found out and \texttt{\textit{fired him}}. He \texttt{\textit{was a bit sad}} but \texttt{\textit{never did}}.
\texttt{\textit{(Conf)}}& \underline{0.14}& 0.62&0.69&0.00\\
\midrule

S5&[FEMALE]'s mom married [FEMALE]'s dad, and the two girls became stepsisters.&[FEMALE], 12, had grown up in a low-income single-parent household. But ani, 7, was wealthy and spoiled, so she was very bratty. At first she hated [FEMALE] and was always mean to her! But then, finally, the two girls began to become friends.&When their dad \texttt{\textit{left the house}}, \texttt{\textit{he went to their room}}. When he came back, he found them in the closet. He scolded them and grounded them for a year. The girls \texttt{\textit{weren't happy with their new stepmother.}}\textit{\texttt{(Chao)}}&\underline{0.00}&0.45&0.52&0.00\\
	\bottomrule
    \end{tabular}
    \caption{Judgments of different metrics for several typical generated samples. \textbf{\underline{H}, M, B} and \textbf{U} stand for human ratings, MoverScore, BLEURT and the proposed metric \textsc{Union}, respectively.  \texttt{\textit{Italic}} words denote the improper entities or events and the specified error type, including \textit{\texttt{Repe}}ated plots, poor \textit{\texttt{Cohe}}rence, \textit{\texttt{Conf}}licting, and \textit{\texttt{Chao}}tic scenes. For comparison, we normalize all the scores to the range of [0,1] following \citeauthor{tao2018ruber}~(\citeyear{tao2018ruber}).}
    \label{tab:case}
  \end{table*}

\begin{table*}[h]
\footnotesize
    \centering
    \begin{tabular}{p{10pt}p{80pt}p{120pt}p{120pt}p{12pt}p{12pt}}
    \toprule
\textbf{ID}&\textbf{Leading Context}&\textbf{Reference}&\textbf{Generated Samples}& \underline{\textbf{H}} &\textbf{\textsc{U}}\\
\toprule
S6&[MALE] wanted to impress his friends.&[MALE] knew how to juggle. [MALE] juggled for his friends. His friends enjoyed the show. [MALE] was happy to captivate his friends attention. &He decided to make a homemade pie for himself. He \texttt{\textit{invited all his friends}} over for a party. Everyone showed up with a recipe. [MALE] was very happy and \texttt{\textit{invited all of his friends. (Rept)}} &\underline{0.43}&0.98\\
\midrule
S7&[FEMALE] came home from school with a lot of homework. & She usually asks her older sister for help. [FEMALE]'s older sister wasn't home. [FEMALE] needed help with her math homework. She then decided to asked her mother for help with her math. & She didn't know what to do with it. She asked her mom to teach her. \texttt{\textit{[FEMALE] said she needed a dog. But [FEMALE]'s mom was a little afraid of him. (Cohe)}}&\underline{0.29}&1.00 \\
\midrule
S8&[FEMALE] was a very well off lawyer in santa fe.&She decided to help a charity in her local area pro bono . [FEMALE] went to the state capital on the charity's behalf. She argued for laws protecting non-profits and the people they serve. [FEMALE] was very successful and felt proud of her accomplishments. & She never had a college degree and had \texttt{\textit{not studied well}}. She \texttt{\textit{was very good at her job}} but always felt like a fraud. A lot of people \texttt{\textit{were disappointed with her}}. [FEMALE] was very upset and decided to major in philosophy.\textit{\texttt{(Conf)}}&\underline{0.14}&1.00\\
\midrule
S9&[MALE] was very nervous.& The big day had finally come and it was time to pop the question. He held her hand, but she didn't know what was going to happen. [MALE] got down on one knee and asked her to marry him. With tears in her eyes she accepted and they embraced.  &He was at the bar with his girlfriend when a man got into his car. He saw that he was going to be alone. The man was \texttt{\textit{shocked by the situation and asked if she was okay}}. [MALE] \texttt{\textit{went to his office and got his wife's name. (Chao)}}&\underline{0.29}&1.00\\
\midrule
S10&[FEMALE] was making coffee before going to work.&But she realized she wouldn't have enough time. So she left her house right away. But when she came back, she realized her stove was still on. So she bought a smoke alarm just in case it happens again. &She didn't put the lid on the pot. She accidentally used the pot on the stove. [FEMALE] burned herself. [FEMALE] is now more careful with her pot. &\underline{0.86}&0.00\\
	\bottomrule
    \end{tabular}
    \caption{Typical misjudgments by \textsc{Union}. \textbf{\underline{H}} and \textbf{U} stand for human ratings and \textsc{Union}, respectively.  \texttt{\textit{Italic}} words denote the improper entities or events and the specified error type, including \textit{\texttt{Repe}}ated plots, poor \textit{\texttt{Cohe}}rence, \textit{\texttt{Conf}}licting, and \textit{\texttt{Chao}}tic scenes.}
    \label{tab:error}
  \end{table*}

\end{document}